\documentclass[lettersize,journal]{IEEEtran}
\usepackage{amsmath,amsfonts}
\usepackage{algorithmic}
\usepackage{algorithm}
\usepackage{array}
\usepackage[caption=false,font=normalsize,labelfont=sf,textfont=sf]{subfig}
\usepackage{textcomp}
\usepackage{stfloats}
\usepackage{url}
\usepackage{verbatim}
\usepackage{graphicx}
\usepackage{cite}
\usepackage{multirow}
\usepackage[table]{xcolor}
\newcommand{\minew}[1]{{\color{black}{#1}}}

\begin{document}

\title{Joint Conditional Diffusion Model for Image Restoration with Mixed Degradations}

\author{Yufeng Yue,~\IEEEmembership{Member,~IEEE,} Meng Yu, Luojie Yang, Yi Yang, ~\IEEEmembership{Member,~IEEE}
\thanks{Manuscript received March 16, 2024.}
\thanks{This work is supported in part by the National Natural Science Foundation of China under Grant 62003039, 62233002; in part by the CAST program under Grant No. YESS20200126.}
\thanks{Yufeng Yue (corresponding author), Meng Yu, and Luojie Yang are with School of Automation, Beijing Institute of Technology, Beijing 100081, China. (e-mail: yueyufeng@bit.edu.cn; 3120225449@bit.edu.cn; 1773259765@qq.com; yang\_yi@bit.edu.cn.)}

}

\markboth{IEEE TRANSACTIONS ON CIRCUITS AND SYSTEMS FOR VIDEO TECHNOLOGY, ~Vol. XXX, No. XXX, March~2024}%
{Yufeng Yue \MakeLowercase{\textit{et al.}}: Joint Conditional Diffusion Model for Image Restoration with Mixed Degradations}


\maketitle

\begin{abstract}
Image restoration is rather challenging in adverse weather conditions, especially when multiple degradations occur simultaneously. Blind image decomposition was proposed to tackle this issue, however, its effectiveness heavily relies on the accurate estimation of each component. Although diffusion-based models exhibit strong generative abilities in image restoration tasks, they may generate irrelevant contents when the degraded images are severely corrupted. To address these issues, we leverage physical constraints to guide the whole restoration process, where a mixed degradation model based on atmosphere scattering model is constructed. Then we formulate our Joint Conditional Diffusion Model (JCDM) by incorporating the degraded image and degradation mask to provide precise guidance. To achieve better color and detail recovery results, we further integrate a refinement network to reconstruct the restored image, where Uncertainty Estimation Block (UEB) is employed to enhance the features. Extensive experiments performed on both multi-weather and weather-specific datasets demonstrate the superiority of our method over state-of-the-art competing methods.
\end{abstract}


\section{Introduction}
\label{sec:intro}

\IEEEPARstart{A}{dverse} weather image restoration is a critical task in computer vision that aims to recover clean images from degraded observations, such as rain, haze, or snow. By enhancing the visual quality of images captured in such conditions, image restoration techniques contribute to improving the accuracy and reliability of subsequent tasks \cite{sun2022rethinking}. 

\begin{figure}[ht!]
\centerline{\includegraphics[width=\columnwidth]{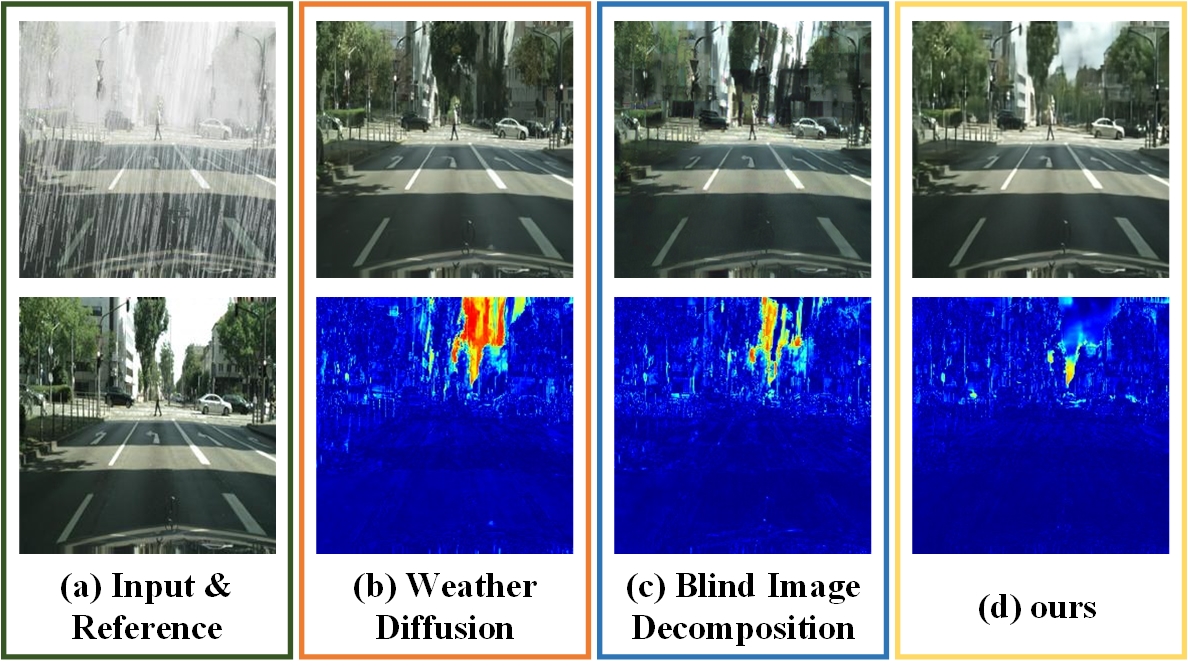}}
\caption{Comparative results of image restoration techniques under the mixed degradations (rain streak + heavy haze). Neither WeatherDiff \cite{ozdenizci2023restoring} nor BIDeN \cite{han2022blind} approaches restore the sky area effectively. The error map highlight the effectiveness of our method in addressing this complex challenge.}
\label{fig1}
\end{figure}

While considerable progress has been made in image restoration for single task, including deraining \cite{ren2019progressive}, \cite{guo2021efficientderain}, \cite{wang2023rcdnet}, dehazing \cite{qin2020ffa}, \cite{guo2022image}, \cite{yu2022frequency}, \cite{song2023vision}, desnowing \cite{liu2018desnownet},\cite{chen2021all}, \cite{zhang2021deep}, \cite{lin2023lmqformer}, and raindrops removal \cite{qian2018attentive}, \cite{shao2021uncertainty}, \cite{iadn}, practical implementation of image restoration faces several challenges. One major obstacle lies in the necessity of correctly identifying the specific degradation type and corruption ratio present in an image. This inconsistency between the prior assumptions made during model construction or training and the unknown degradation hampers the effectiveness of these methods. To release this assumption, researchers have turned their attention to develop an all-in-one model \cite{li2020all}, \cite{valanarasu2022transweather}, \cite{Cui2023IRNeXtRC} to handle multiple weather degradations. Notably, the All-in-one approach \cite{li2020all} was the first unified model to cope with three weather types by employing separate encoders, while TransWeather \cite{valanarasu2022transweather} introduced
a single shared encoder. \minew{Considering the different frequency characteristics, AIRFormer \cite{airformer} further designed an frequency-oriented transformer encoder and decoder.} Despite their generic architecture, these methods were primarily applied to recover one specific task. 

Nevertheless, in complex and dynamic environments, the degradation can change rapidly and even occur simultaneously, which poses a considerable challenge for all-in-one approaches. For instance, a heavy rain image may exhibit both rain streaks and haze caused by the rain veiling effect. Built upon this, AirNet \cite{li2022all} was presented to handle combined degradations involving rain and haze by analyzing the inherent characteristics. To extend its applicability to a wider range of possible degradation types, Han \textit{et al}. \cite{han2022blind} reconsidered image restoration with mixed degradation as a Blind Image Decomposition (BID) problem and used a CNN-based network to separate these components without knowing the degradation type. However, this network necessitates laborious training due to the employment of multiple decoders for each component, including the degradation masks. Meanwhile, an accurate estimation of the degradation mask is crucial for successful decomposition, otherwise it will affect the effectiveness of the recovery task, seen in Fig. 1. Motivated by this, we intend to develop techniques that can effectively handle complex and diverse degradation scenarios, without the need to explicitly identify or separating individual degradation component. 

Recently, the successful applications of diffusion model \cite{ho2020denoising} in various image restoration tasks \cite{croitoru2023diffusion} have demonstrated their stronger expressiveness in learning the underlying data distribution. Researchers have extended these models to address adverse weather degradation removal, such as dehazing \cite{dehazeddpm} and deraining \cite{raindiffusion}. More recently, WeatherDiff \cite{ozdenizci2023restoring} initially presented a generic model for multi-weather restoration tasks. Although these diffusion-based methods can achieve high-resolution restoration, they may produce innovative content unrelated to the original image due to their stronger generative capabilities. This is particularly evident when dealing with severely degraded images, they often suffer from significant information loss in terms of texture and fine details. In addition, relying solely on the degraded image as a condition may not provide sufficient guidance. Motivated by these observations, we aim to introduce physical constraints to guide the generative process. Based on the atmosphere scattering model, we construct the mixed degradation model, so that the model can receive various types of degraded images. Furthermore, by incorporating the degradation binary mask, the diffusion model can focus more on the specific degraded regions of the image. We further integrate a refinement network to enhance the restoration process.

In summary, the main novelty of this paper is to design a diffusion-based image restoration model that can effectively handle the challenges posed by mixed weather degradations. For this paper, the main contributions are as follows:
\begin{enumerate}
    \item A mixed weather degradation model based on the atmospheric scattering model is constructed, which can be regarded as a foundational model to generate combined weather degradation.
    \item We propose a novel Joint Conditional Diffusion Model (JCDM), which introduces degraded image and predicted mask as conditions to guide the restoration process. 
    \item In the refinement restoration stage, the Uncertainty Estimation Block (UEB) is utilized to enhance the color and detail recovery.
\end{enumerate}

The rest of this paper is organized as follows. Section II describes recent related works. Section III demonstrates the proposed methodology. Section IV analyzes the qualitative and quantitative experiments and results on various datasets. Finally, Section V concludes our work.

\section{Related Work}
\label{sec:2}
In this section, we will provide a concise overview of recent advancements in image restoration and discuss the relevant methods that are addressed in this paper.

\subsection{Single Image Restoration with Specific Degradation}
In recent years, there have been remarkable advancements in the field of single image restoration. Existing image restoration methods can be mainly categorized into independent and all-in-one approaches.

\subsubsection{Independent Image Restoration Methods}
Various techniques have been developed to address specific types of weather degradation, such as rain streaks, raindrop, haze, and snow, based on the premise that there is only one type of degradation present. These techniques, including deraining \cite{ren2019progressive, jiang2020multi, purohit2021spatially, guo2021efficientderain, wang2023rcdnet}, dehazing \cite{qin2020ffa, guo2022image, dehazeddpm, yu2022frequency, song2023vision}, and desnowing \cite{liu2018desnownet, chen2020jstasr, chen2021all, zhang2021deep, lin2023lmqformer}, leverage separate networks for task-specific training. Although several existing methods \cite{zamir2021multi, zamir2022restormer, cui2023focal} have been proposed as general restoration networks, they still require tedious one-by-one training and fine-tuning on individual datasets. Moreover, one notable requirement of these methods is the accurate selection of specific degradation types and levels corresponding to different environmental conditions.

\subsubsection{All-in-one Image Restoration Methods}
Some researchers have investigated the use of a single model to address multiple weather removal problems. The All-in-One approach \cite{li2020all} was initially proposed to tackle various weather degradations within a unified model, employing task-specific encoders and a shared decoder. To alleviate the computational complexity, Transweather \cite{valanarasu2022transweather} introduced a single-encoder single-decoder network based on vision transformer \cite{dosovitskiy2020image}, incorporating learnable specific weather queries to handle general weather degradation conditions, while IRNeXt \cite{Cui2023IRNeXtRC} performed filter modulation on the attention weights to accentuate the informative spectral part of feature. \minew{Based on the different frequency characteristics observed in the early and late stages of feature extraction, AIRFormer \cite{airformer} introduced a frequency-guided Transformer encoder and a frequency-refined Transformer decoder.} Additionally, \cite{chen2022learning} combined the two-stage knowledge learning strategy and multi-contrastive knowledge regularization loss to tackle a specific adverse weather removal problem. Similarly, \cite{zhu2023learning} designed a two-stage network to explore the weather-general and weather-specific features separately, allowing for adaptive expansion of specific parameters at learned network positions. However, it is crucial to highlight that these methods are limited to dealing with a single degradation each time.

\subsection{Blind Image Restoration with Mixed Degradation}
Notably, the corrupted images captured in adverse weather conditions often exhibit a combination of multiple degradations. For instance, on a rainy day, a degraded image may contain raindrops, haze, and other forms of degradation. Recognizing the limitations of single image restoration methods designed for specific tasks, researchers began to explore more efficient and robust image restoration frameworks that do not rely on prior knowledge of the degradation.

To address multiple types of degradations simultaneously, Li \textit{et al}. \cite{li2022all} presented AirNet in an all-in-one fashion, effectively capturing the inherent characteristics of combined degradations involving rain and haze. Han \textit{et al}. \cite{han2022blind} further proposed the BID setting, treating degraded images as arbitrary combinations of individual components such as rain streaks, snow, haze, and raindrops. However, this network employed multiple decoders for each decomposed component, which required tedious training procedures. Expanding on these advancements, we aim to design a model that can effectively restore degraded images with complex degradations without the need to explicitly identify or separate individual degradation components. 


\subsection{Diffusion-based Image Restoration}
Recently, Denoising Diffusion Probabilistic Models (DDPM) \cite{ho2020denoising} have achieved remarkable success in a wide range of image restoration tasks with higher quality \cite{croitoru2023diffusion}. Based on the foundational diffusion models, researchers have extended these models to address single weather degradation removal, such as dehazing \cite{dehazeddpm} and deraining \cite{raindiffusion}. For instance, DehazeDDPM \cite{dehazeddpm} combined atmosphere scattering model with DDPM, incorporating the separated physical components into the denoising process. While this approach achieved high-resolution recovery, it generated scenario-independent information that may not be optimal for all situations. More recently, WeatherDiff \cite{ozdenizci2023restoring} enhanced the capabilities of diffusion models for handling multiple weather conditions. It introduced a patch-based image restoration algorithm for arbitrary sized image processing. However, such diffusion models treat degraded images as guided conditions for the restoration process. While when faced with severe weather degradation, the condition may be too weak to provide more useful information, leading to creative image generation. In contrast to the aforementioned approaches, we aim for introducing physical constraints in the conditional diffusion models to guide and enhance the restoration process, enabling effective blind image restoration in challenging scenarios.

\section{Proposed Method}
\label{sec:3}
In this section, we describe and formulate the algorithm, divided into five subsections: Architecture Design, Physical Mixed Degradation Model, Joint Conditional Diffusion Model, Refinement Restoration, and Loss Function.

\begin{figure*}[ht!]
\centerline{\includegraphics[width=0.95\textwidth]{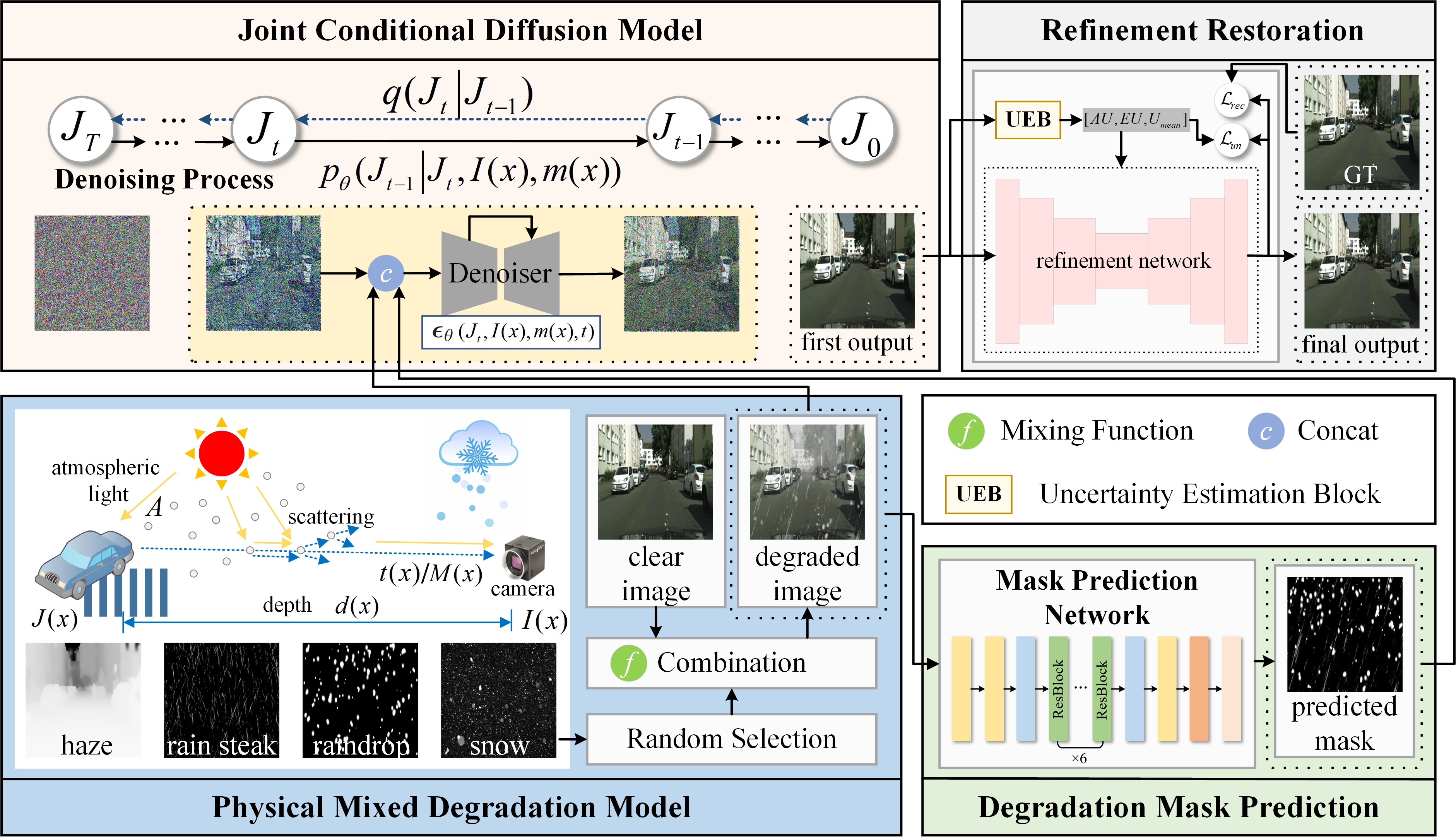}}
\caption{Overall architecture of the proposed algorithm. The pipeline is as follows. Firstly, we formulate the degraded image with random mixed degradations using our constructed model (equation (8)). Subsequently, the mask prediction branch is leveraged to estimate the degradation mask corresponding to the degraded image. This predicted mask, along with the degraded image, serves as conditions of the diffusion model. The initial restoration results obtained from the stage are then fed into the refinement network. Finally, the restored image is obtained.}
\label{fig2}
\end{figure*}

\subsection{Architecture Design}

To address the formulated problem of blind image restoration, we adopt a two-step approach, as depicted in Fig. 2. Building upon the principles of conditional diffusion models, we consider the degraded image, constructed by equation (8), as one of the conditions for the restoration process. Additionally, we leverage the information provided by the degradation mask, which indicates the location and size of the corrupted areas, as another condition. Here, we extend the capabilities of the methodology proposed in the previous work \cite{raindropmask} to predict the degradation mask. By employing the Joint Conditional Diffusion Model, we generate a coarse output that serves as an initial restoration. Finally, we design a refinement network that incorporates an Uncertainty Estimation Block (UEB) to effectively restrain the uncertainty in the restoration process and achieve high-quality image restoration results.

\subsection{Physical Mixed Degradation Model}
Under adverse weather conditions, the captured image can be corrupted by various types of degradation. The most popular rain streaks model used in existing studies is the additive composite model \cite{li2016rain}, which can be expressed as:
\begin{equation}
I(x)=J(x)+\sum\limits_{t=1}^{s}S_t.
\label{eq1}
\end{equation}
where $I(x)$ represents the \emph{x}-th pixel of the observed degraded image, and $J(x)$ is the corresponding clear image. $S_t$ denotes the rain-streak layer that has the same streak direction. The index $t$ represents the rain-streak layer and $s$ is the maximum number of the rain-streak layers. 

Moreover, according to \cite{qian2018attentive}, an adherent raindrop image is modelled as :
\begin{equation}
I(x)=J(x)\odot(1-M_d(x))+R(x).
\label{eq2}
\end{equation}
where $\odot$ denotes element-wise multiplication, $M_d(x)$ is the binary mask and $R(x)$ is the imagery brought about by the adherent raindrops, representing the blurred imagery formed the light reflected by the environment. 

Generally, a snowy image can be modelled as\cite{liu2018desnownet}:
\begin{equation}
I(x)=J(x)\odot(1-M_s(x))+A\odot M_s(x).
\label{eq3}
\end{equation}
where $M_s(x)$ is the binary mask, indicating the location of snow. In the mask, $M_s(x)=1$ means the pixel $x$ is part of a snow region, and otherwise means it is part of background regions.

Lastly, based on the atmospheric scattering theory \cite{ASM}, the hazy image can be mathematically modelled as follows:   
\begin{equation}
I(x)=J(x)\odot t(x)+A\odot(1-t(x)).
\label{eq4}
\end{equation}
\begin{equation}
t(x)=e^{-\beta d(x)}.
\label{eq5}
\end{equation}
where $t(x)$ denotes the transmission map, which is exponentially correlated to scene depth $d(x)$ and scattering coefficient $\beta$ that reflects the haze density.

The above physical degradation models are applicable to the modeling of a single degradation. When multiple degradations occur simultaneously, for example, a heavy rain image may contain rain streaks and fog/mist caused by rain veiling effect, the single degradation model may be difficult to characterize the various weather type. According to the atmospheric scattering model, the atmospheric light value will decrease correspondingly in adverse weather conditions. Therefore, we can further reconstruct the rain degradation model combined with the atmospheric light.

As for the rain streaks model, we leverage a mask $M_r(x)$ to represent the rain streaks, then the equation (\ref{eq1}) can be further modelled as:
\begin{equation}
I(x)=J(x)\odot(1-M_r(x))+A\odot M_r(x).
\label{eq6}
\end{equation}

As for of raindrop image modelling, $R(x)$ in can be further expressed as $R(x)=A\odot M_d(x)$. Then the equation (\ref{eq2}) can be written as:
\begin{equation}
I(x)=J(x)\odot(1-M_d(x))+A\odot M_d(x).
\label{eq7}
\end{equation}

Then, a degradation model with random mixed multiple degradations is proposed, which can be represented as follows:
\begin{equation}
I(x)=\mathcal{G}^n(\mathcal{T}((J(x),t(x)),M(x)).
\label{eq8}
\end{equation}
where $M(x)$ represents the degradation binary mask, including $M_r(x)$, $M_d(x)$, and $M_s(x)$. $n=0,1,2,3$ means the number of degradation types. The function $\mathcal{G}(\cdot,\cdot)$ and $\mathcal{T}(\cdot,\cdot)$ are reflection functions, defined as:
\begin{equation}
\mathcal{G}(a,b)=a\odot (1-b)+A\odot b.
\label{eq9}
\end{equation}
\begin{equation}
\mathcal{T}(a,b)=a\odot b+A\odot (1-b).
\label{eq10}
\end{equation}
where $a$ and $b$ are input images, and the function combines them with the global atmosphere light $A$.


\subsection{Joint Conditional Diffusion Model}

Diffusion models\cite{ho2020denoising},\cite{song2020denoising} are generative models which are aimed at learning the process of converting a Gaussian distribution into the targeted data distribution. Diffusion models generally can be divided into forward diffusion process and reverse diffusion process.

The forward process, which is inspired by non-equilibrium thermodynamics\cite{sohl2015deep}, can be viewed as a fixed Markov Chain to corrupt initial image $J(x)$ by gradually adding noise according to a variance schedule $\beta_1,...,\beta_T$, where the initial data distribution can be regarded as $J_0 \sim q(J_0)$. After $T$ time steps of sequentially adding noise, the obtained data distribution $J_T \sim q(J_T)$ is nearly a normal distribution and the forward diffusion process can be modelled as:
\begin{equation}
    q(J_t|J_{t-1}) = \mathcal N(J_t;\sqrt{1-\beta_t} J_{t-1},\beta_t I),
\label{eq11}
\end{equation}
\begin{equation}
    q(J_{1:T}|J_0) = \prod\limits_{t=1}\limits^{T}{q(J_t|J_{t-1})}.
\label{eq12}
\end{equation}

For the forward process, it is notable that sampling arbitrary latent variables $J_1,..,J_T$ in closed form is admitted according to the equation (11) and equation (12) by using the notation $\alpha_t=1-\beta_t$ and $\bar{\alpha}_t=\prod\limits_{s=1}\limits^{t}{\alpha_s}$ and it can be formulated as:
\begin{equation}
    q(J_t|J_0) = \mathcal N(J_t;\sqrt{\bar{\alpha}_t} J_0,(1-\bar{\alpha_t}) I).
\label{eq13}
\end{equation}

The reverse diffusion process, which reverses the forward process, can recreate desired data distribution through gradually denoising transitions starting from prior $q(J_T)=\mathcal N(J_T;0, I)$. Due to the difficulties to estimate $q(J_{t-1}|J_t)$, the conditional probabilities are approximated by a learned model $p_\theta$ with KL divergences. The approximate joint distribution $p_\theta(J_{0:T})$ can be mathematically modelled as follows:
\begin{equation}
    p_\theta(J_{0:T}) = p(J_T)\prod\limits_{t=1}\limits^{T}{p_\theta(J_{t-1}|J_{t})},
\label{eq14}
\end{equation}
\begin{equation}
    p_\theta (J_{t-1}|J_{t}) = \mathcal N(J_{t-1};\mu_\theta(J_t,t),\Sigma_{\theta}(J_t,t)).
\label{eq15}
\end{equation}

As conditional diffusion models \cite{rombach2022high},\cite{chung2022come} have displayed outstanding capabilities of data editing, a conditional denoising diffusion process $p_\theta(J_{0:T}|I(x),m(x))$ is learned to preserve more features from the degraded image $I(x)$ and better remove weather interference by information of the predicted mask $m(x)$. Converted from the equation (14), the conditional denoising diffusion process can be represented as:
\begin{equation}
    p_\theta(J_{0:T}|I(x),m(x)) = p(J_T)\prod\limits_{t=1}\limits^{T}{p_\theta(J_{t-1}|J_{t},I(x),m(x))},
\label{eq16}
\end{equation}

\begin{algorithm}[ht!]
\caption{Joint conditional diffusion model training}
\textbf{Input:} Initial clear image $J(x)$, denoted as $J_0$, corresponding degraded image $I(x)$, the time step $T$, and the degradation binary mask $m(x)$.
\begin{algorithmic}[1] 
    \REPEAT
        \STATE $J_0\sim q(J_0)$
        \STATE $t\sim $Uniform$({1,...,T})$
        \STATE $\epsilon_t \sim \mathcal N(0,I)$
        \STATE Perform a single gradient descent step for   
        
          \qquad $\nabla_\theta ||\epsilon_t-\epsilon_\theta(\sqrt{\bar{\alpha}_t}J_0+\sqrt{1-\bar{\alpha}_t}\epsilon_t,I(x),m(x),t)||^2$ 
    \UNTIL converged
\STATE return $\theta$
\end{algorithmic}
\end{algorithm}

\begin{algorithm}[ht!]
\caption{Joint degradation diffusive image restoration}
\textbf{Input:} Degraded image $I(x)$, conditional diffusion model $\epsilon_\theta(J_t,I(x),m(x),t)$, the time step $T$, the number of implicit sampling steps $S$, and the degradation binary mask $m(x)$.
\begin{algorithmic}[1] 
    \STATE $J_T \sim \mathcal N(0,I)$
    \FOR{ $i=S:1$ }
        \STATE$t=(i-1)\cdot T/S + 1$
        \STATE$t_{next}=(i-2)\cdot T/S + 1$ \textbf{if} $i>1$ \textbf{else} $0$
        \STATE$J_{t-1}\leftarrow\sqrt{\bar{\alpha}_{t_{next}}}(\frac{J_t-\sqrt{1-\bar{\alpha}_t} \cdot \epsilon_\theta(J_t,I(x),m(x),t)}{\sqrt{\bar{\alpha}_t}}) + \sqrt{1-\bar{\alpha}_{t_{next}}} \cdot \epsilon_\theta(J_t,I(x),m(x),t)$
    \ENDFOR
\STATE return $J_0$
\end{algorithmic}
\end{algorithm}

To guide conditional denoising transitions to an expected output, the training process is conducted by optimizing the function approximator  $\epsilon_\theta(J_t,I(x),m(x),t)$, through which $\mu_\theta$ can be gotten, and stochastic gradient descent is applied to adjust $\epsilon_\theta$. The joint degradation conditional diffusion model training process is summarized in Algorithm 1. The objective function can be modelled as:
\begin{equation}
    L = E_{J_0,\epsilon_t \sim \mathcal N(0,I),t} \big[||\epsilon_t-\epsilon_\theta(J_t,I(x),m(x),t)||^2\big].
\label{eq17}
\end{equation}

The joint degradation diffusive image restoration process is summarized in Algorithm 2. The image restoration process through reverse diffusive sampling is deterministic and accelerated with an implicit denoising process \cite{ho2020denoising}, of which the formulation is as follows:
\begin{equation}
\begin{split}
    J_{t-1}=\sqrt{\bar{\alpha}_{t_{next}}}(\frac{J_t-\sqrt{1-\bar{\alpha}_t} \cdot \epsilon_\theta(J_t,I(x),m(x),t)}{\sqrt{\bar{\alpha}_t}})\\
    + \sqrt{1-\bar{\alpha}_{t_{next}}} \cdot \epsilon_\theta(J_t,I(x),m(x),t),
\end{split}   
\label{eq18}
\end{equation}

\subsection{Refinement Restoration}

\begin{figure}[ht!]
\centerline{\includegraphics[width=0.9\columnwidth]{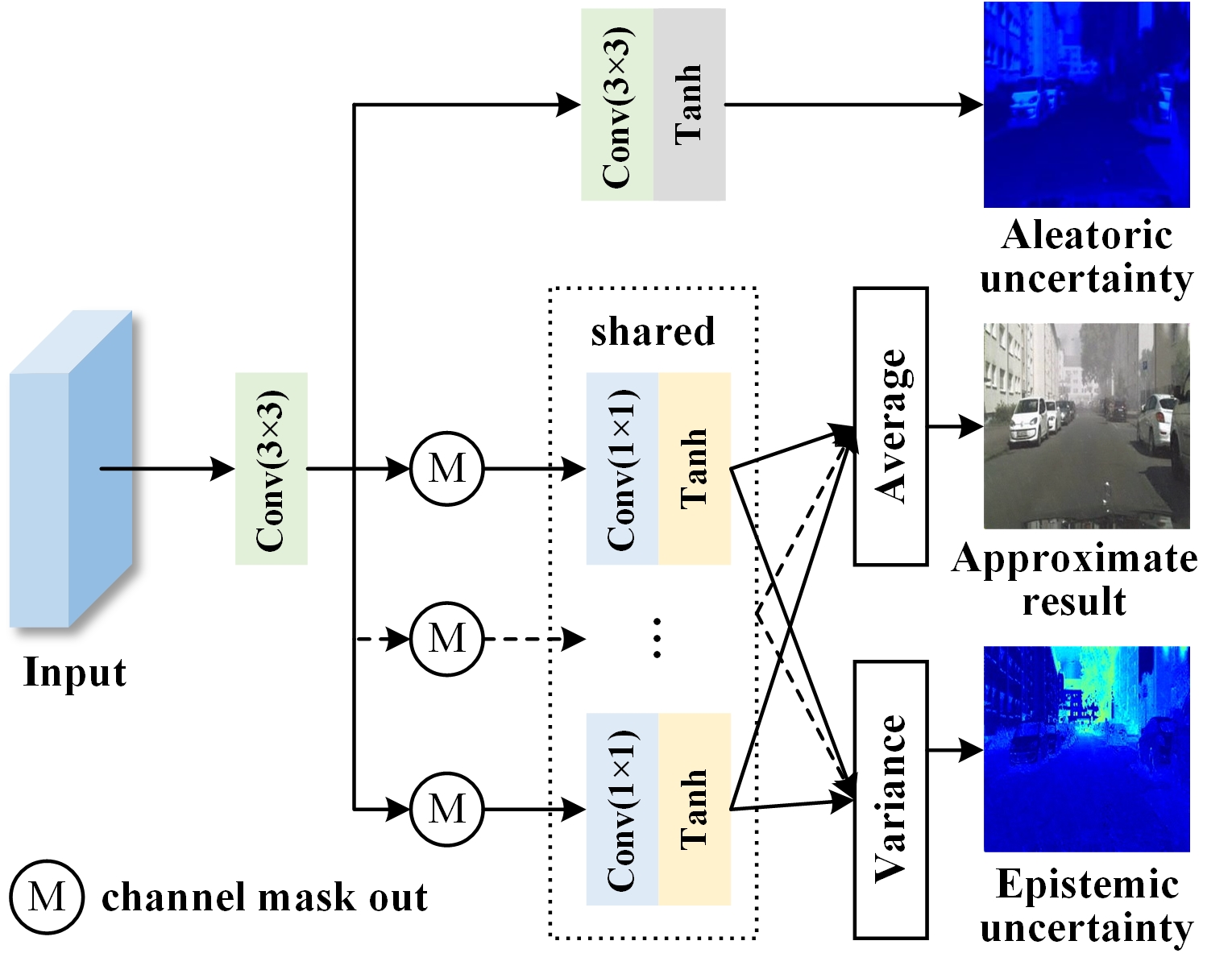}}
\caption{The overall Uncertainty Estimation Block (UEB) structure, which conducts aleatoric and epistemic uncertainty modeling through two separate branches, respectively.}
\label{fig3}
\end{figure}


After the first step of recovery, we utilize a refinement network to restore more details and achieve high resolution, where a U-shaped network is employed to explore abundant features at each scale. In the network, UEB is incorporated to enhance the dependable features. Specifically, according to kendall \textit{et al}. \cite{kendall2017uncertainties}, two kinds of uncertainty arise in deep learning models. One is epistemic uncertainty, which can describe the uncertainty of the model's predictions, while another is associated with the inherent noise present in the observations, called aleatoric uncertainty. Inspired by previous work \cite{uncertainty}, the UEB is introduced to model each pixel's epistemic uncertainty and aleatoric uncertainty, as shown in Fig. 3. 

In the UEB framework, we begin by feeding the input feature into a convolution layer, then the output is split into two separate branches. The upper branch is responsible for estimating the aleatoric uncertainty $U_A$. It further processes the feature map through an additional convolution layer followed by a Sigmoid function. On the other hand, the bottom branch involves sampling the input feature multiple times $S_T$. In each sampling, we randomly mask a certain percentage $q$ of the channels. This random masking procedure introduces diversity and helps capture different potential representations of the input. Each of the sampling results $J_a$ then undergoes a shared convolution layer and a Tanh activation function. By averaging the results of these sampling operations, we obtain the mean prediction, serves as an approximate restoration result. While the epistemic uncertainty $U_E$ is obtained by calculating the variance. Then the predicted uncertainty of each pixel can be approximated as the following expression.
\begin{equation}
U\approx U_E+U_A
\label{eq19}
\end{equation}



Then, we can leverage the UEB to enhance the refinement network and improve the restoration process. In detail, during the feature extraction stage at the \textit{i}-th scale, assuming the input feature is denoted as ${\mathbf{F}_{in}^i}$, and the extracted feature is ${\mathbf{F}_{out}^i}$, we can estimate the uncertainty map $U_i$ of the input feature using UEB. Subsequently, the modulated feature $\mathbf{F}_m^i$ at the \textit{i}-th scale can be mathematically modelled as:
\begin{equation}
\mathbf{F}_m^i=\mathbf{F}_{in}^i \odot U_i+{\mathbf{F}_{out}^i}\odot (1-U_i).
\label{eq20}
\end{equation}

To sum up, with this modulation operation, the final restoration result, denoted as $J_f(x)$, is obtained. Fig. 4 provides a visual comparison of the restoration results with and without refinement. In areas where uncertainty is initially high, the refinement network has successfully reduced the uncertainty and improved the quality of the restored image.

\begin{figure}[ht!]
\centerline{\includegraphics[width=0.9\columnwidth]{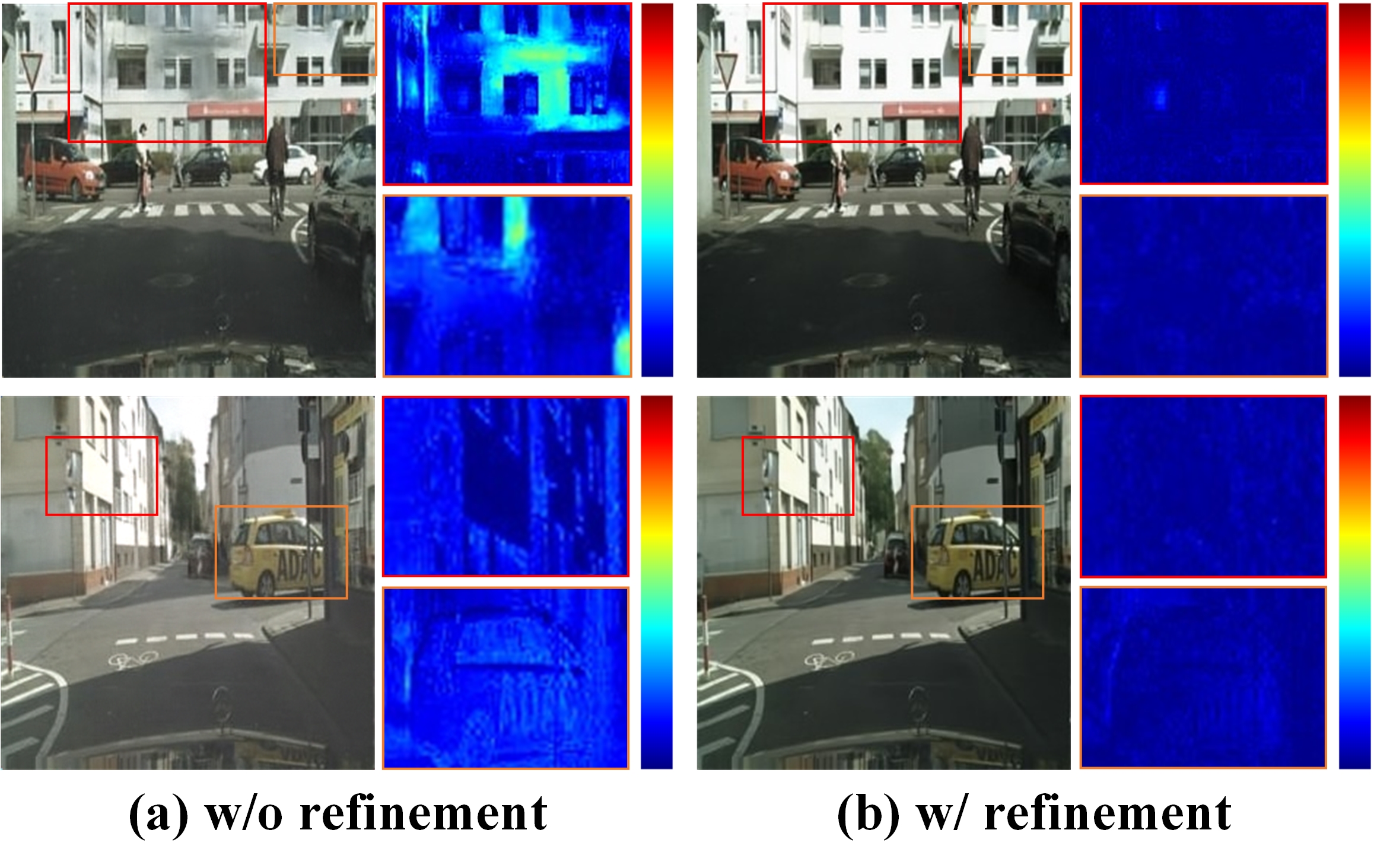}}
\caption{Comparison of the restoration results with and without refinement.}
\label{fig4}
\end{figure}

\subsection{Loss Function}
The total loss function $\mathcal{L}_{all}$ designed for model optimization consists of two parts: reconstruction loss $\mathcal{L}_{rec}$ and uncertainty-aware loss $\mathcal{L}_{un}$, which is formulated as follows.
\begin{equation}
\mathcal{L}_{all}=\mathcal{L}_{rec}+\lambda \mathcal{L}_{un}.
\label{eq21}
\end{equation}
where $\lambda$ is the corresponding coefficient.

Particularly, the reconstruction loss $\mathcal{L}_{rec}$ is obtained by calculating the Mean squared error (MSE) (L1 loss) between the clear image $J(x)$ and final restoration image $J_f(x)$, which is defined as:
\begin{equation}
\mathcal{L}_{rec}=\Vert J(x)-J_f(x) \Vert _1.
\label{eq22}
\end{equation}

Moreover, the uncertainty-aware loss $\mathcal{L}_{un}$ can be represented as the following expression.
\begin{equation}
\mathcal{L}_{un}=\Vert J(x)-J_{a}(x) \Vert _1+\mathcal{L}_{au}.
\label{eq23}
\end{equation}
where $\mathcal{L}_{au}$ is formulated in the uncertainty estimation process, which can be modelled as follows.
\begin{equation}
\mathcal{L}_{au}=\dfrac{1}{N}\sum_{j=1}^N(\alpha e^{-{U_{A}^j}}(I^j(x)-J_{a}^j(x))^2+\beta U_{A}^j).
\label{eq24}
\end{equation}
where $N$ denotes the number of pixels, $\alpha$ and $\beta$ represent the weighting factors.

\section{Experiments}
\label{sec:4}
In this section, extensive experiments are conducted to validate the effectiveness of our proposed method. In the following, we will first introduce the experimental settings and then present the qualitative and quantitative comparison results with state-of-the-art baseline methods. Finally, we will conduct several ablation studies.

\subsection{Experimental Settings}
\subsubsection{Restoration Tasks} 
We carry out two sub-tasks to evaluate the restoration performance. \textbf{Task I}: Joint degradation (raindrop/rainstreak/snow/haze) removal, \textbf{Task II}: Specific degradation removal. 

\subsubsection{Datasets} 
For joint degradation removal, following the constructed mixed degradation model, we generate the corrupted images with random mixed combinations based on the CityScape \cite{cordts2016cityscapes} dataset. The masks for rain streak and snow are acquired from Rain100H \cite{yang2017deep} and Snow100K \cite{liu2018desnownet}, while the raindrop masks adopt the metaball model \cite{blinn1982generalization} to model the droplet shape and property with various random locations, numbers and sizes.

For specific degradation removal, we use four standard benchmark image restoration datasets considering adverse weather conditions of rain, haze, and snow. For image deraining, Raindrop dataset \cite{qian2018attentive} consists of real adherent-raindrop images for raindrop removal. For image dehazing, Dense-Haze \cite{ancuti2019dense} and NH-HAZE \cite{ancuti2020nh} datasets are introduced with the NTIRE Dehazing Challenges, which show different haze densities according to local image areas. For image desnowing, Snow100K \cite{liu2018desnownet} is a dataset for evaluation of image desnowing models, which comprises three Snow100K-S/M/L sub-test sets, indicating the synthetic snow strength imposed via snowflake sizes (light/mid/heavy). 

\subsubsection{Comparison Baseline}
To verify the effectiveness of our proposed method, we compare it with several representative and state-of-the-art baseline methods. 

For joint degradation removal, we divide all the baselines into 3 groups, consisting of task-agnostic methods (\textit{i}.\textit{e}., MPRNet \cite{zamir2021multi}, Restormer\cite{zamir2022restormer}, FocalNet \cite{cui2023focal}), multi-task-in-one methods (\textit{i}.\textit{e}., All-in-one \cite{li2020all}, TransWeather \cite{valanarasu2022transweather}, IRNeXt \cite{Cui2023IRNeXtRC}, WeatherDiff \cite{ozdenizci2023restoring}), and blind IR method (\textit{i}.\textit{e}., BIDeN \cite{han2022blind}). Among this, the task-agnostic methods have a unified scheme for different tasks but need to be trained separately, multi-task-in-one methods can remove different types of weather using a single set of parameters. The first two groups are designed for specific degradation and blind IR methods are designed for mixed degradation.


\begin{table*}[ht]
\caption{Quantitative results of joint degradation removal. We evaluate the performance in PSNR and SSIM under 6 cases, which are (1) rain streak, (2) rain streak + snow, (3) rain streak + light haze, (4) rain streak + heavy haze, (5) rain streak + moderate haze + raindrop, (6) rain streak + snow + moderate haze + raindrop. The best performance under each case is marked in \textbf{bold} with the second performance \underline{underlined}.}
\setlength{\tabcolsep}{5pt}
\renewcommand{\arraystretch}{1.2}
\centering
\begin{tabular}{|c|c|cc|cc|cc|cc|cc|cc|} 
\hline
\multirow{2}{*}{Type}   & \multirow{2}{*}{Method} & \multicolumn{2}{c|}{case 1} & \multicolumn{2}{c|}{case 2} & \multicolumn{2}{c|}{case 3} & \multicolumn{2}{c|}{case 4} & \multicolumn{2}{c|}{case 5} & \multicolumn{2}{c|}{case 6}  \\ 
\cline{3-14}
 &     & PSNR  & SSIM    & PSNR  & SSIM    & PSNR  & SSIM      & PSNR  & SSIM  & PSNR  & SSIM    & PSNR  & SSIM       \\ 
\hline
\multirow{3}{*}{task-agnostic} 
& MPRNet \cite{zamir2021multi}   & 33.95 & 0.945  & 30.52 & 0.909    & 23.98 & 0.900   & 18.54 & 0.829   & 21.18 & 0.846  & 20.76 & 0.812    \\  
& Restormer \cite{zamir2022restormer}   & 34.29 & 0.951    & 30.60 & 0.917     & 23.74 & 0.905  & 20.33 & 0.853  & 22.17 & 0.859  & 21.24 & 0.821  \\    
& FocalNet \cite{cui2023focal}    & 34.27 & 0.955    & 29.00 & 0.894           & 27.64 & 0.911 & 27.80 & 0.913  & 26.35 & 0.888  & 24.70 & 0.834    \\ 
\hline
\multirow{4}{*}{multi-task in one}    
& All-in-one \cite{li2020all}  & 32.38 & 0.937   & 28.45 & 0.892               & 27.14 & 0.911   & 19.67 & 0.865  & 24.23 & 0.889   & 22.93 & 0.846    \\
& TransWeather \cite{valanarasu2022transweather}    & 27.91 & 0.833            & 26.29 & 0.770 & 24.93 & 0.792   & 24.99 & 0.794   & 23.51 & 0.756            & 23.03 & 0.716   \\
& IRNeXt \cite{Cui2023IRNeXtRC}   & 35.18 & 0.957  & 30.25 & 0.901            & \underline{28.94} & \underline{0.920}   & \underline{29.05} & \textbf{0.922}   & \underline{27.63} & 0.901  & 26.18 & 0.855    \\
& WeatherDiff \cite{ozdenizci2023restoring}        & 35.13 & 0.943             & 29.93 & 0.869  & 28.45 & 0.893   & 28.54 & 0.896    & 27.31 & 0.871          & 25.64 & 0.811                \\ 
\hline
\multirow{3}{*}{blind IR}                                                 
& BIDeN \cite{han2022blind}  & 30.89 & 0.932  & 29.34 & 0.899  & 28.62 & 0.919 & 26.77 & 0.891   & 27.11 & 0.898      & 26.44 & 0.870      \\
& \cellcolor{gray!20}ours (w/o refinement)    & \cellcolor{gray!20}\underline{35.44}&\cellcolor{gray!20}\underline{0.962}
&\cellcolor{gray!20}\underline{32.93}&\cellcolor{gray!20}\underline{0.937} 
&\cellcolor{gray!20}28.76   & \cellcolor{gray!20}0.916                   &\cellcolor{gray!20}28.66   & \cellcolor{gray!20}\underline{0.916}             &\cellcolor{gray!20}27.39   &\cellcolor{gray!20}\underline{0.910}              &\cellcolor{gray!20}\underline{27.51}  &\cellcolor{gray!20}\underline{0.880} \\
& \cellcolor{gray!20}ours (w/ refinement)    
& \cellcolor{gray!20}\textbf{38.34}   & \cellcolor{gray!20}\textbf{0.966}     & \cellcolor{gray!20}\textbf{34.07}   & \cellcolor{gray!20}\textbf{0.945}     & \cellcolor{gray!20}\textbf{29.84}   & \cellcolor{gray!20}\textbf{0.928}     & \cellcolor{gray!20}\textbf{29.40}   & \cellcolor{gray!20}\textbf{0.922}
&\cellcolor{gray!20}\textbf{27.73}    & \cellcolor{gray!20}\textbf{0.912}
&\cellcolor{gray!20}\textbf{27.86}    & \cellcolor{gray!20}\textbf{0.885}   \\
\hline

\end{tabular}
\label{tab1}
\end{table*}

\begin{figure*}[ht!]
\centerline{\includegraphics[width=\textwidth]{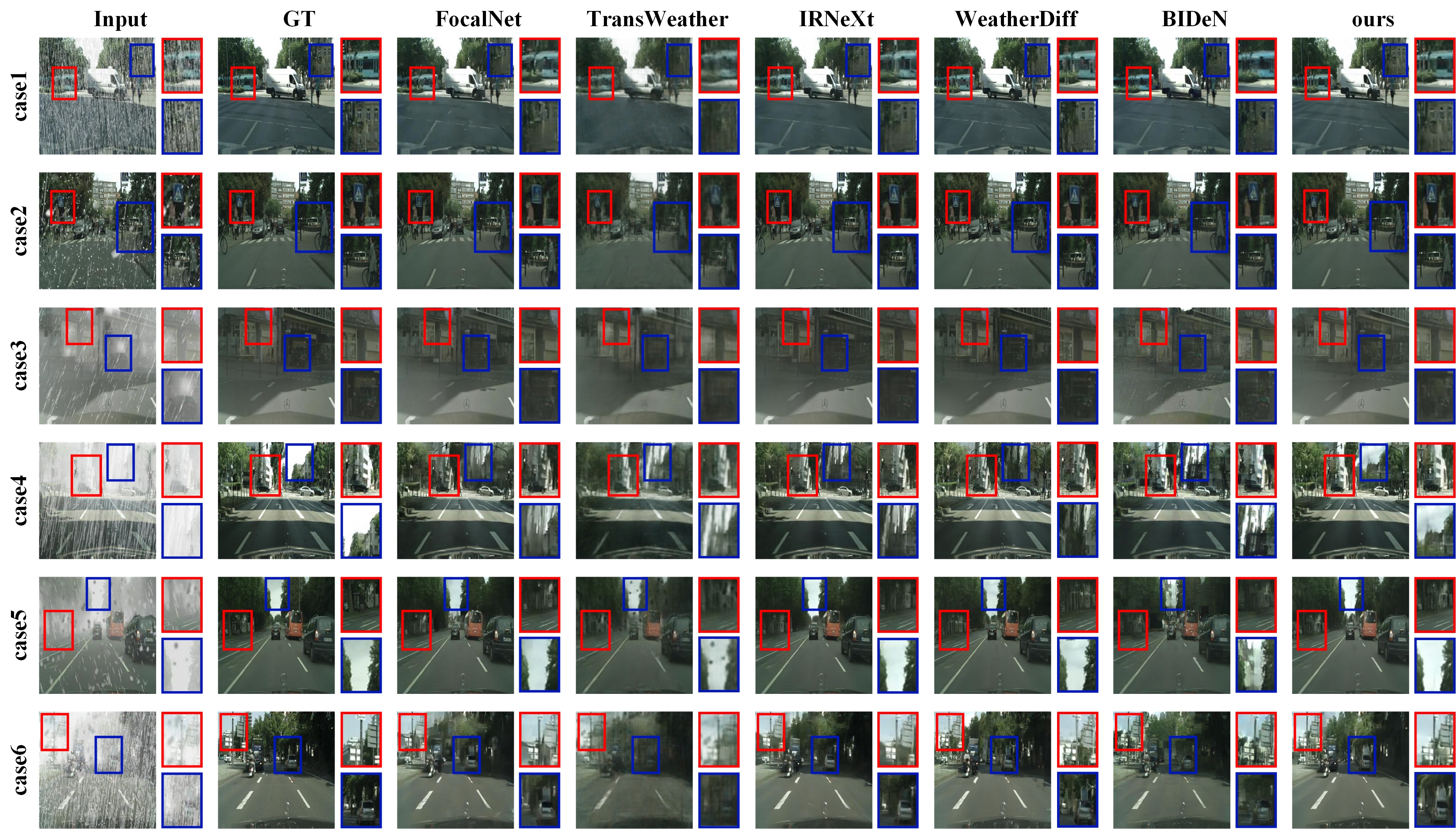}}
\caption{Qualitative results of joint degradation removal under 6 cases. Some areas are highlighted in colored rectangles for a better visualization and comparison.}
\label{fig5}
\end{figure*}

For specific degradation removal, apart from the above task-agnostic and multi-task-in-one methods, we supplement the task-specific methods designed for a specific kind of weather. In terms of image deraining, we compare with PreNet \cite{ren2019progressive}, \minew{IADN \cite{iadn}} and EfficientDerain \cite{guo2021efficientderain}, while for image dehazing, we compare with FFANet \cite{qin2020ffa}, Dehamer \cite{song2023vision}, FSDGN \cite{yu2022frequency}, Dehazeformer \cite{guo2022image}, and \minew{SDBAD-Net \cite{sdbad}}. Moreover, methods including DesnowNet \cite{liu2018desnownet}, HDCW-Net \cite{chen2021all}, \minew{DesnowGAN \cite{DesnowGAN}}, DDMSNET \cite{zhang2021deep}, and LMQFormer \cite{lin2023lmqformer} are compared for image desnowing.

\subsubsection{Implementation Details}
All experiments are conducted on a desktop system with an NVIDIA Geforce RTX 4090 GPU. We use Adam optimizer with the momentum as (0.9, 0.999) for optimization. The batch size and patch size are set to 8 and $256 \times256$, respectively. Moreover, the initial learning rate is set as $3\times 10^{\mbox{-}5}$. During the training phase, 1000 diffusion steps were performed, while the noise schedule $\beta_t$ linearly increased from 0.0001 to 0.02. For inference, a total of 25 steps were utilized.

\subsubsection{Evaluation Metrics}
We adopt two popular metrics for quantitative comparisons, including Peak Signal-to-Noise Ratio (PSNR) and Structure Similarity Index (SSIM). Higher value of these metrics indicates better performance of the image restoration methods.

\subsection{Qualitative and Quantitative Analysis}

\subsubsection{Joint degradation Removal}
Table I and Fig. 5 provide a comprehensive comparison between our method and several baselines. Notably, task-agnostic methods exhibit strong performance in case 1, where there is a single type of degradation. However, their performance significantly deteriorates in more complex situations, indicating their limited robustness in handling diverse and mixed weather degradations. Among the multi-task in one methods, except for TransWeather, the other three methods demonstrate better performance in more complex cases. The All-in-one method benefits from its multi-head encoder, enabling it to learn more universal features, while the generative capabilities of WeatherDiff contribute to its improved performance.

\begin{table}[t]
\caption{Quantitative results of image deraining on Raindrop dataset. The best performance under each case is marked in \textbf{bold} with the second performance \underline{underlined}.}
\setlength{\tabcolsep}{4pt}
\renewcommand{\arraystretch}{1.2}
\centering
\begin{tabular}{|c|c|cc|} 
\hline
\multirow{3}{*}{Type}   & \multirow{3}{*}{Method} & \multicolumn{2}{c|}{image deraining}  \\ 
\cline{3-4} &  & \multicolumn{2}{c|}{Raindrop}     \\ 
\cline{3-4} &  & PSNR & SSIM   \\ 
\hline
\multirow{3}{*}{task-specific}                                            
& PreNet \cite{ren2019progressive} & 24.96 & 0.863  \\
& \minew{IADN \cite{iadn}}  & \minew{25.65}  & \minew{0.824}  \\
& EfficientDerain \cite{guo2021efficientderain}  & 28.48  & 0.897  \\
\hline
\multirow{3}{*}{task-agnostic} 
& MPRNet \cite{zamir2021multi}& 28.33    & 0.906  \\
& Restormer \cite{zamir2022restormer} & 28.32   & 0.888  \\
& FocalNet \cite{cui2023focal}  & 25.09    & 0.913  \\ 
\hline
\multirow{5}{*}{multi-task in one}                                        
& All-in-one \cite{li2020all}    & 31.12 & 0.927  \\
& TransWeather \cite{valanarasu2022transweather} & 28.84   & 0.843  \\
& IRNeXt \cite{Cui2023IRNeXtRC} & 24.63    & 0.902  \\
& WeatherDiff \cite{ozdenizci2023restoring} & 30.71  & 0.931  \\
& \minew{AIRFormer \cite{airformer}} & \minew{30.04}  & \minew{0.948}  \\
\hline
\multirow{2}{*}{blind IR}                                                  
& \cellcolor{gray!20}ours (w/o refinement)
&\cellcolor{gray!20}\underline{31.57}    &\cellcolor{gray!20}\underline{0.935}   \\
& \cellcolor{gray!20}ours (w/ refinement) 
&\cellcolor{gray!20}\textbf{31.82} 
&\cellcolor{gray!20}\textbf{0.939}      \\
\hline
\end{tabular}
\end{table}

\begin{figure*}
\centerline{\includegraphics[width=\textwidth]{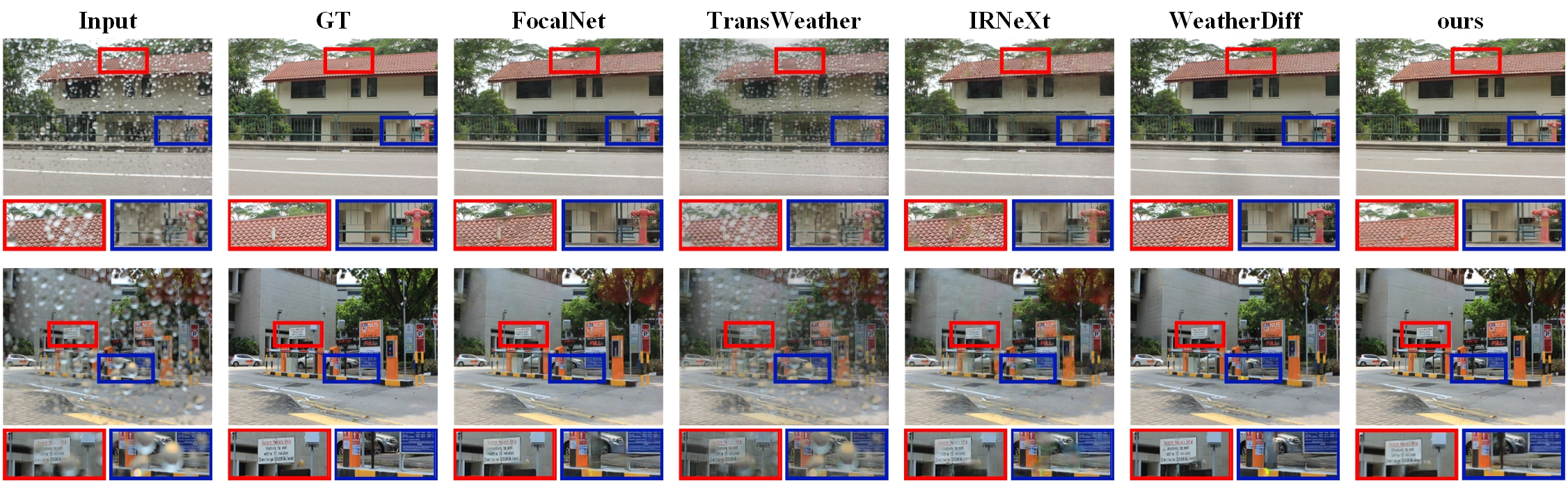}}
\caption{Qualitative results of raindrop removal on Raindrop dataset. Some areas are highlighted in colored rectangles for a better visualization and comparison.}
\label{fig7}
\end{figure*}

It is important to note that our proposed method achieves competitive performance across all cases. Although it may slightly underperform compared to IRNeXt in case 3, 4 and 5, the inclusion of our refinement network consistently enhances restoration results across various scenarios. Specifically, as a diffusion-based method, our approach outperforms WeatherDiff in all cases, with or without refinement, and shows remarkable restoration quality in fine details (enlarged in red and blue bounding boxes). When compared with blind image restoration methods, despite that BIDeN is able to handle more complex cases, its performance in simple scenarios like case 1 and case 2 is limited by its training setting and insufficient feature learning. Furthermore, our method maintains higher generality under complex conditions. For example, the PSNR of 27.86 in case 6 exceeds that of 27.73 in case 5.

\subsubsection{Image Deraining}
Table II provides the quantitative results of image deraining task on Raindrop dataset. Our method achieves the best metrics in terms of PSNR and SSIM, indicating its superior performance compared to other methods. Additionally, Fig. 6 showcase visualizations of image deraining reconstructions for sample test images. Our method demonstrates the ability to restore cleaner images that closely resemble the ground truth. In particular, our method excels in preserving details in areas sheltered by raindrops. In contrast, seen in the highlighted rectangles, competing methods may erase these details (\textit{i}.\textit{e}., WeatherDiff, TransWeather) or introduce artifacts (\textit{i}.\textit{e}., FocalNet, IRNeXt). The visual performance of our method stands out by restoring more accurate and detailed representations.

\begin{table}[t]
\caption{Quantitative results of image dehazing on Dense-Haze and NH-HAZE dataset. The best performance under each case is marked in \textbf{bold} with the second performance \underline{underlined}.}
\setlength{\tabcolsep}{3.5pt}
\renewcommand{\arraystretch}{1.2}
\centering
\begin{tabular}{|c|c|cc|cc|} 
\hline
\multirow{3}{*}{Type}   & \multirow{3}{*}{Method} & \multicolumn{4}{c|}{image dehazing}  \\ 
\cline{3-6}
&  & \multicolumn{2}{c|}{Dense-Haze} & \multicolumn{2}{c|}{NH-HAZE}     \\ 
\cline{3-6}
&    & PSNR  & SSIM          & \multicolumn{1}{c}{PSNR} & SSIM   \\ 
\hline
\multirow{4}{*}{task-specific}                                            
& FFANet \cite{qin2020ffa}    & 14.39 & 0.452  & 19.87 & 0.692  \\
& Dehamer \cite{song2023vision}    & 16.62 & 0.560  & 20.66  & 0.684  \\
& FSDGN \cite{yu2022frequency}  & 16.91 & 0.581  & 19.99  & 0.711      \\
& Dehazeformer \cite{guo2022image}   & 16.29 & 0.510   & 20.47  & 0.731  \\
& \minew{SDBAD-Net \cite{sdbad}}   & - & -  & \minew{19.89}  & \minew{0.743}  \\
\hline
\multirow{3}{*}{task-agnostic} 
& MPRNet \cite{zamir2021multi}  & 15.36 & 0.574   & 19.27  & 0.675  \\
& Restormer \cite{zamir2022restormer} & 15.72 & 0.619  & 19.60 & 0.704  \\
& FocalNet \cite{cui2023focal} & 17.07 & \underline{0.630} & 20.43 & \textbf{0.790} \\ 
\hline
\multirow{3}{*}{multi-task in one}                                        
& TransWeather \cite{valanarasu2022transweather} &12.44 &0.349 &14.52&0.269\\
& WeatherDiff \cite{ozdenizci2023restoring}& 12.28 & 0.472 & 13.66 & 0.537 \\
& \minew{AIRFormer \cite{airformer}}& - & - & \minew{\underline{20.85}} & \minew{0.740} \\
\hline
\multirow{2}{*}{blind IR}                                               
&\cellcolor{gray!20}ours (w/o refinement)      &\cellcolor{gray!20}16.74    &\cellcolor{gray!20}0.599    &\cellcolor{gray!20}19.32    &\cellcolor{gray!20}0.705        \\
& \cellcolor{gray!20}ours (w/ refinement)   &\cellcolor{gray!20}\textbf{17.56}   &\cellcolor{gray!20}\textbf{0.635}    &\cellcolor{gray!20}\textbf{20.87}    &\cellcolor{gray!20}\underline{0.755}        \\
\hline
\end{tabular}
\end{table}

\subsubsection{Image Dehazing}
Table III presents the quantitative results of image dehazing task on Dense-Haze and NH-HAZE datasets. The results show that our method ranks in the front second among all the methods evaluated. In comparison to the task-agnostic method FocalNet, although our method has a lower SSIM, the higher PSNR metric suggests that our diffusion model contributes to restoring more realistic structures. \minew{Compared with transformer-based methods such as Dehamer, Restormer, and AIRFormer, our diffusion-based method demonstrates superior performance due to its generative capability in invisible regions.} Furthermore, Fig. 7 and Fig. 8 depict visualizations of the image dehazing results on the above two datasets, respectively. Our method outperforms other methods in terms of haze removal and the preservation of structural details. These visual results further emphasize the superior performance of our method in improving visual quality and removing haze artifacts. Overall, our approach achieves competitive quantitative metrics and produces visually pleasing results, showcasing its capabilities in haze removal.

\subsubsection{Image Desnowing}
Table IV and Fig. 9 present a thorough comparison between our method and several baselines on the Snow100K-L dataset. With the inclusion of refinement, our method achieves notable improvements in terms of RSNR and SSIM. Specifically, the refined results exhibit an improvement of 0.44 in RSNR and 0.009 in SSIM, indicating that our method achieves SOTA performance among all the methods compared. When compared to IRNeXt, FocalNet, and Weatherdiffusion, our method stands out in terms of restoration quality, particularly in fine details as highlighted in the enlarged red and blue bounding boxes in Fig. 9. 

\begin{figure*}[t]
\centerline{\includegraphics[width=\textwidth]{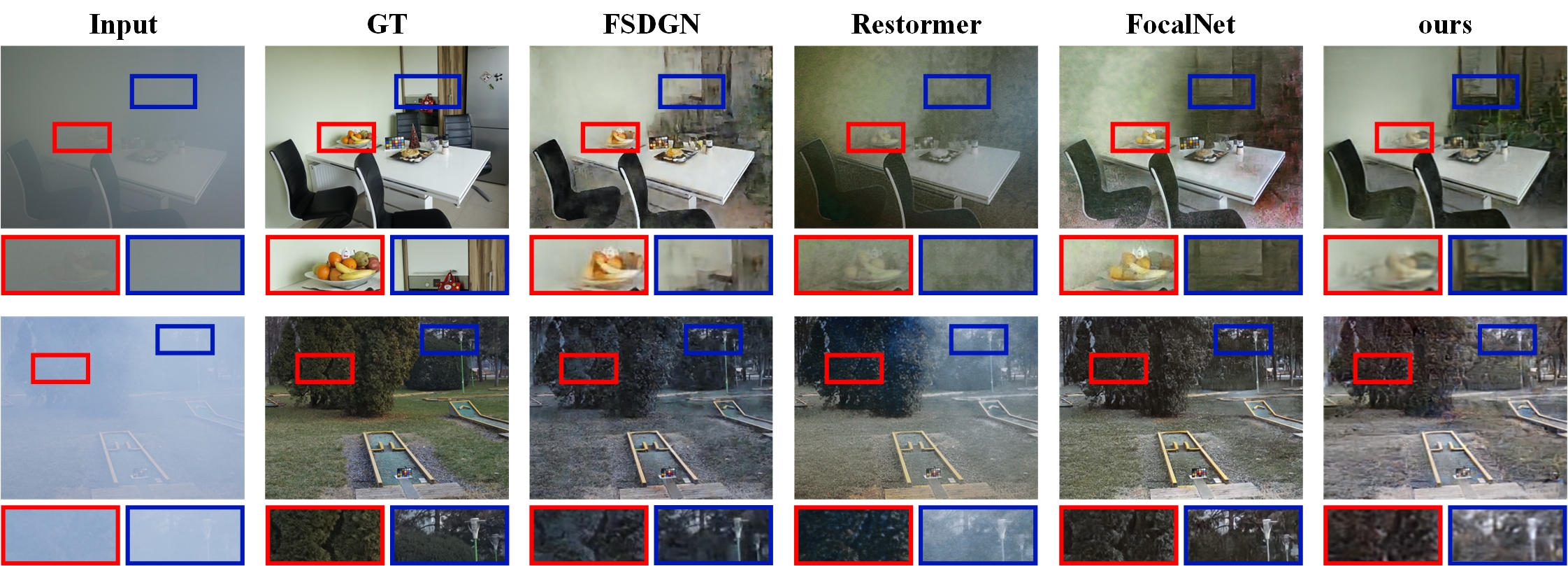}}
\caption{Qualitative results of image dehazing on Dense-Haze dataset. Some areas are highlighted in colored rectangles for a better visualization and comparison.}
\label{fig8}
\end{figure*}

\begin{figure*}[t]
\centerline{\includegraphics[width=\textwidth]{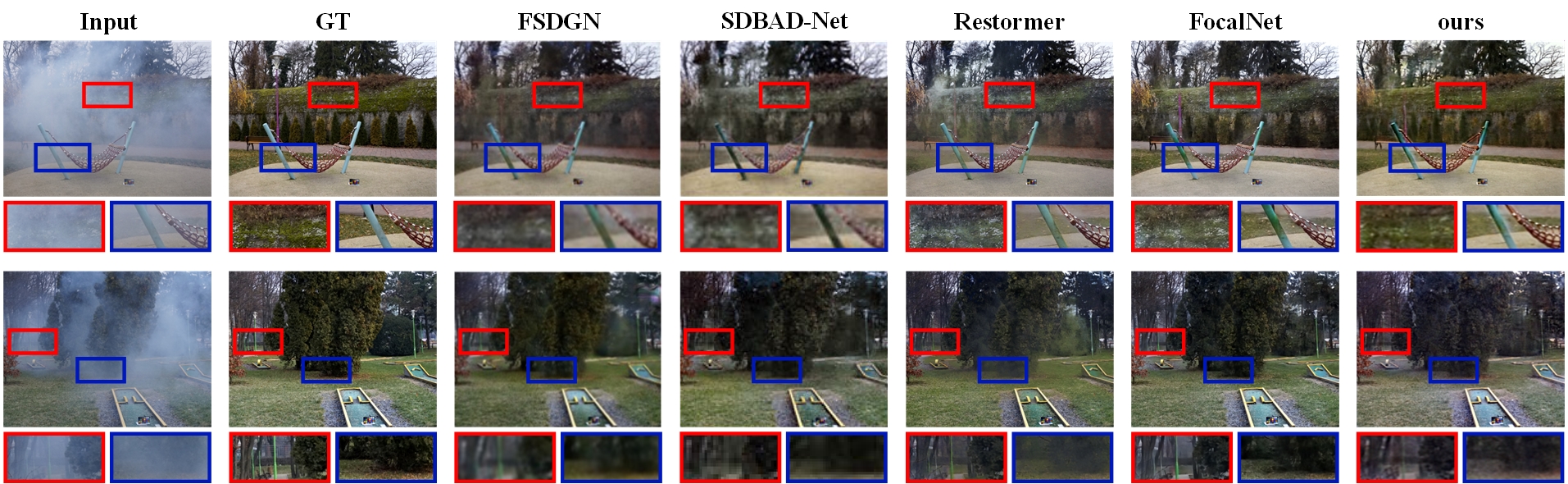}}
\caption{\minew{Qualitative results of image dehazing on NH-HAZE dataset. Some areas are highlighted in colored rectangles for a better visualization and comparison.}}
\label{fig9}
\end{figure*}

\begin{table}
\caption{Quantitative results of image desnowing on Snow100K-L dataset. The best performance under each case is marked in \textbf{bold} with the second performance \underline{underlined}.}
\renewcommand{\arraystretch}{1.2}
\centering
\begin{tabular}{|c|c|cc|} 
\hline
\multirow{3}{*}{Type} & \multirow{3}{*}{Method} & \multicolumn{2}{c|}{image desnowing}  \\ 
\cline{3-4}&    & \multicolumn{2}{c|}{Snow100K-L}  \\ 
\cline{3-4}&  & PSNR  & SSIM    \\ 
\hline
\multirow{5}{*}{task-specific}                                            
& DesnowNet \cite{liu2018desnownet}  & 27.17 & 0.898    \\
& HDCW-Net \cite{chen2021all}        & 20.88 & 0.618    \\
& \minew{DesnowGAN \cite{DesnowGAN}}   & \minew{28.07} & \minew{0.921}    \\
& DDMSNET \cite{zhang2021deep}       & 28.85 & 0.877    \\
& LMQFormer \cite{lin2023lmqformer}  & 29.71 & 0.890    \\ 
\hline
\multirow{3}{*}{task-agnostic} 
& MPRNet \cite{zamir2021multi}        & 29.76 & 0.895        \\
& Restormer \cite{zamir2022restormer} & \underline{30.83} & 0.912   \\
& FocalNet \cite{cui2023focal}        & 30.15 & \underline{0.927}   \\ 
\hline
\multirow{5}{*}{multi-task in one}                                        
& All-in-one \cite{li2020all}       & 28.33 & 0.882      \\
& TransWeather \cite{valanarasu2022transweather}    & 29.31 & 0.888      \\
& IRNeXt \cite{Cui2023IRNeXtRC}          & 30.81 & \textbf{0.929}      \\
& WeatherDiff \cite{ozdenizci2023restoring}     & 30.09 & 0.904      \\ 
& \minew{AIRFormer \cite{airformer}} & \minew{29.00}  & \minew{0.925}  \\
\hline
\multirow{2}{*}{blind IR} 
& \cellcolor{gray!20}ours (w/o refinement)  &\cellcolor{gray!20}30.64       &\cellcolor{gray!20}0.920       \\
& \cellcolor{gray!20}ours (w/ refinement)   &\cellcolor{gray!20}\textbf{31.08}       &\cellcolor{gray!20}\textbf{0.929}       \\
\hline
\end{tabular}
\end{table}

\subsection{Ablation Study}
\subsubsection{Effect of Joint Conditional Diffusion Model}
In our conditional diffusion model, we leverage the degradation mask and degraded image as the conditions. To verify its effectiveness in improving the image quality during the restoration process, we conduct experiments on Task I, specifically focusing across case 4 to case 6. In our experiments, we adopt the conditional denoising diffusion probabilistic model as the baseline, which only considers the degraded image as the condition. Building upon this, our method introduce the degradation mask as joint condition, and the findings are reported in Table V. The results clearly demonstrate that incorporating the degradation mask as a conditional factor leads to a significant improvement in image restoration performance. This improvement is particularly noticeable when dealing with cases involving complex combinations of degradations.

\begin{table}
\caption{Ablation study on the effect of joint conditional diffusion model.}
\setlength{\tabcolsep}{4pt}
\renewcommand{\arraystretch}{1.2}
\centering
\begin{tabular}{|c|cc|cc|cc|} 
\hline
\multirow{2}{*}{Method} & \multicolumn{2}{c|}{case 4} & \multicolumn{2}{c|}{case 5} & \multicolumn{2}{c|}{case 6}  \\ 
\cline{2-7} & PSNR & SSIM & PSNR & SSIM  & PSNR & SSIM     \\ 
\hline
baseline  & 22.91  & 0.898 & 21.34 & 0.887 & 19.48 & 0.784 \\
joint condition  & \textbf{29.40} & \textbf{0.922} & \textbf{27.73} & \textbf{0.912} & \textbf{27.86} & \textbf{0.885} \\
\textbf{\textit{improvement}} & \textit{+6.49} & \textit{+0.024} & \textit{+6.39} & \textit{+0.025} & \textit{+8.38} & \textit{+0.101} \\
\hline
\end{tabular}
\end{table}

\subsubsection{Effect of Refinement Network}
The refinement network is an integral component that follows the initial restoration process and aims to further enhance the quality and fidelity of the restored images. In this study, we compare the performance of our full approach with two variations: one where the refinement network was not utilized (referred to as “w/o refinement”), and another where the refinement network was employed (referred to as “w/ refinement”). Comparing the restoration results between the scenarios with and without refinement, we can see notable enhancements across various evaluation metrics, presented in Table I to IV. Furthermore, Fig. 4 provides a visual comparison that further supports the improved image restoration performance achieved with the inclusion of the refinement network. Specifically, the images restored with the refinement network showcase sharper edges, lower uncertainty, and better color representation.

\begin{figure*}[t]
\centerline{\includegraphics[width=\textwidth]{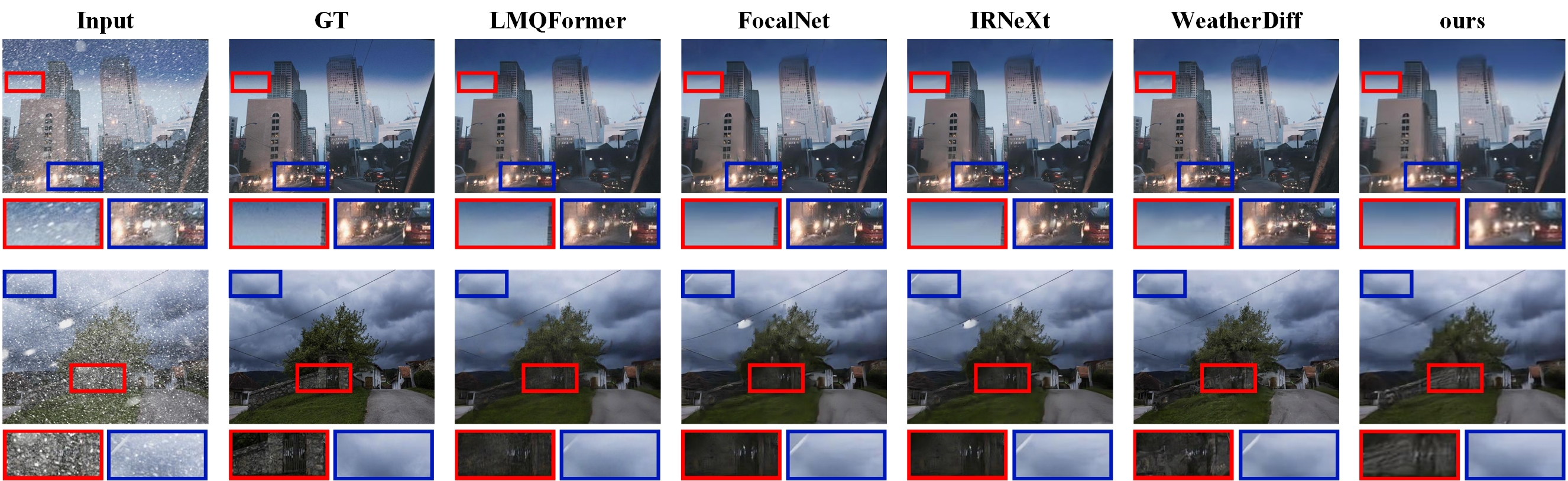}}
\caption{Qualitative results of image desnowing on Snow100K-L dataset. Some areas are highlighted in colored rectangles for a better visualization and comparison.}
\label{fig10}
\end{figure*}

\subsection{Efficiency}
Table VI presents a comparison of the number of parameters, FLOPs (floating-point operations), and inference time efficiency among several competitive methods. The reported time corresponds to the average inference time for each model using test images of dimensions $256\times256$, ensuring a fair comparison. Our method demonstrates superior inference time efficiency compared to the diffusion-based method WeatherDiff. It is over 100 times faster, indicating a significant improvement in computational efficiency. Despite the increased speed, our method maintains competitive performance, achieving superior results in terms of restoration quality. This combination of faster inference time and superior performance makes our method a compelling choice for time-sensitive applications.

\begin{table}
\caption{Discussion on the model efficiency. All models are tested under the same environment for fair comparisons.}
\renewcommand{\arraystretch}{1.2}
\centering
\begin{tabular}{|c|ccc|} 
\hline
Method & Para (M) & FLOPs (G) & Inference time (s)  \\ 
\hline
FocalNet \cite{cui2023focal} & 3.74  & 30.53  & 0.0064    \\
TransWeather \cite{valanarasu2022transweather} & 37.68 & 6.13 & 0.218 \\
IRNeXt \cite{Cui2023IRNeXtRC}  & 5.45  & 41.95  & 0.0133  \\
WeatherDiff \cite{ozdenizci2023restoring} & 82.92 & 475.16 & 22.253 \\
BIDeN \cite{han2022blind}  & 39.812 & 214.15 & 0.149  \\ 
\minew{AIRFormer \cite{airformer}} & \minew{58.34} & \minew{5.746} & \minew{0.086} \\
\hline
\rowcolor{gray!20} ours (w/o refinement) & 55.49 & 182.06 & 0.194 \\
\rowcolor{gray!20} ours (w/ refinement)  & 61.82 & 225.39 & 0.216 \\
\hline
\end{tabular}
\end{table}

\section{Conclusion}
This paper proposes a diffusion-based method for image restoration, specifically targeting adverse weather conditions. The proposed approach addresses the challenge of combined degradations by incorporating the degraded image and corresponding degradation mask as conditional information. This inclusion enables more targeted and adaptive restoration, leading to improved image quality and accuracy. Additionally, a refinement network is integrated to enhance the initial restoration results. Experimental results demonstrate the significant performance improvement achieved through our approach, especially in complex scenarios. Moving forward, future research efforts can focus on refining the diffusion process to better preserve semantic details while effectively removing degradation. 

\section*{Acknowledgments}
This work is supported in part by the National Natural Science Foundation of China under Grant 62003039, 62233002; in part by the CAST program under Grant No. YESS20200126.

\bibliographystyle{IEEEtran}
\bibliography{reference.bib}


\section{Biography Section}

\begin{IEEEbiography}[{\includegraphics[width=1in,height=1.25in,clip,keepaspectratio]{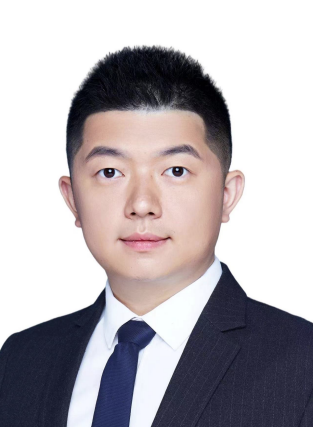}}]{Yufeng Yue} (Member, IEEE) received the B.Eng. degree in automation from the Beijing Institute of Technology, Beijing, China, in 2014, and the Ph.D. degree in electrical and electronic engineering from Nanyang Technological University, Singapore, in 2019. He is currently a Professor with School of Automation, Beijing Institute of Technology. He has published a book in Springer, and more than 60 journal/conference papers, including IEEE TMM/TMech/TII/TITS, and conferences like NeurIPS/ICCV/ICRA/IROS. He is an Associate Editor for 2020–2024 IEEE IROS. His research interests include perception, mapping and navigation for autonomous robotics.
\end{IEEEbiography}

\vspace{11pt}

\begin{IEEEbiography}[{\includegraphics[width=1in,height=1.25in,clip,keepaspectratio]{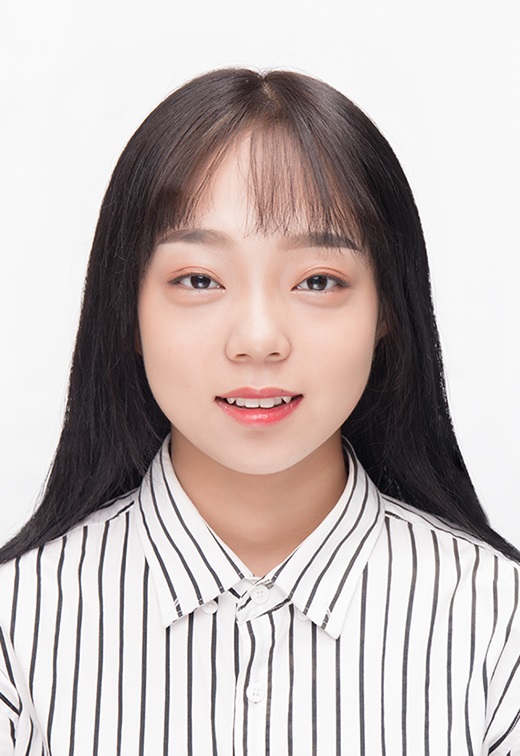}}]{Meng Yu} received the B.Eng. degree in automation from the School of Instrument and Electronics, North University of China, in 2020. She is currently a Ph.D student in Control Science and Engineering with the School of Automation, Beijing Institute of Technology. Her research interests include multimodal sensor fusion, robotics perception, and computer vision.
\end{IEEEbiography}

\vspace{11pt}

\begin{IEEEbiography}[{\includegraphics[width=1in,height=1.25in,clip,keepaspectratio]{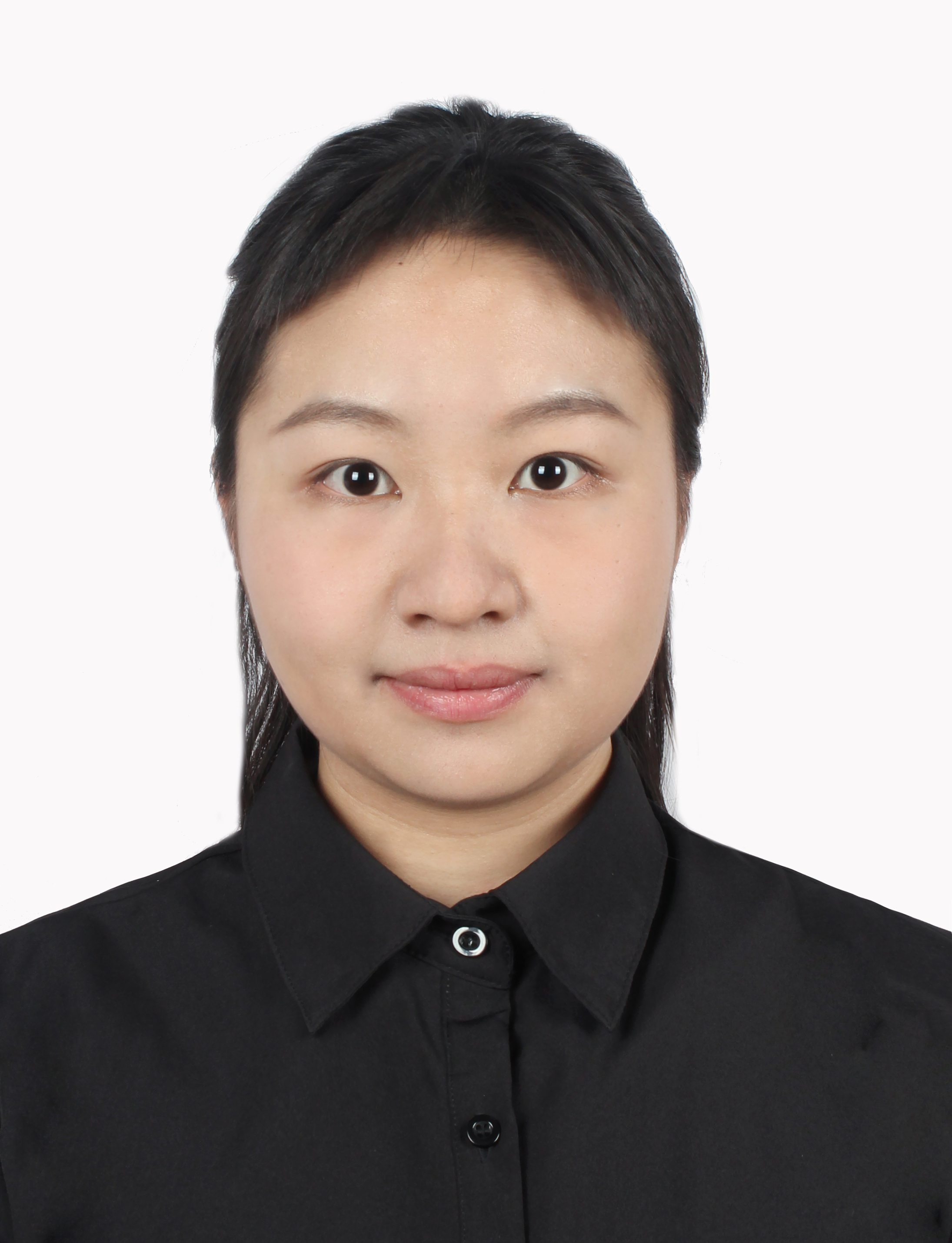}}]{Luojie Yang} is currently an undergraduate student at the School of Automation, Beijing Institute of Technology. She will be continuing her studies as a Master’s student in Navigation, Guidance and Control with the School of Automation, Beijing Institute of Technology. Her research interests include computer vision, multimodal sensor fusion and robotics perception.
\end{IEEEbiography}

\vspace{11pt}

\begin{IEEEbiography}[{\includegraphics[width=1in,height=1.25in,clip,keepaspectratio]{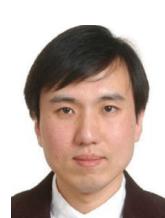}}]{Yi Yang} received the B.Eng. and M.Eng. degrees from the Hebei University of Technology, Tianjin, China, in 2001 and 2004, respectively, and the Ph.D. degree from the Beijing Institute of Technology, Beijing, China, in 2010. He is currently a Professor with the School of
Automation, Beijing Institute of Technology. His research interests include robotics, autonomous systems, intelligent navigation, cross-domain collaborative perception, and motion planning and control. Dr. Yang received the National Science and Technology Progress Award twice.
\end{IEEEbiography}



\vfill

\end{document}